\pdfoutput=1

\documentclass[11pt]{article}
\usepackage{CJKutf8}
\DeclareUnicodeCharacter{2212}{-}
\usepackage{booktabs}
\usepackage[]{acl}
\usepackage{graphicx} 
\usepackage{times}
\usepackage{latexsym}
\usepackage{algorithm}

\usepackage{algpseudocode}
\usepackage{bm}
\usepackage{tabularx}
\usepackage{lipsum}
\usepackage{color,xcolor}
\usepackage{booktabs}

\usepackage[T1]{fontenc}
\usepackage[T5]{fontenc}

\usepackage[utf8]{inputenc}

\usepackage{microtype}

\title{Mitigating Catastrophic Forgetting in Multi-domain Chinese Spelling Correction by Multi-stage Knowledge Transfer Framework}

\author{
Peng Xing$^{1}$\thanks{\ \ indicates equal contribution.},~Yinghui Li$^{1*}$,~Shirong Ma$^{1}$,~Xinnian Liang$^{2}$,~Haojing Huang$^{1}$\\~\textbf{Yangning Li}$^{1}$,~\textbf{Hai-Tao Zheng}$^{1}$\thanks{\ \ Corresponding authors (cswhjiang@gmail.com, zheng.haitao@sz.tsinghua.edu.cn).},~\textbf{Wenhao Jiang}$^{3\dagger}$,~\textbf{Ying Shen}$^{4}$\\
        $^{1}$ Tsinghua Shenzhen International Graduate School, Tsinghua University \\ 
        $^{2}$ State Key Lab of Software Development Environment, Beihang University \\
        $^{3}$ Guangdong Laboratory of Artificial Intelligence and Digital Economy (SZ) \\
        $^{4}$ School of Intelligent Systems Engineering, Sun-Yat Sen University\\
        \texttt{\{xp23, liyinghu20\}@mails.tsinghua.edu.cn}
}

\begin{document}
\maketitle
\begin{abstract}
Chinese Spelling Correction (CSC) aims to detect and correct spelling errors in given sentences. 
Recently, multi-domain CSC has gradually attracted the attention of researchers because it is more practicable.
In this paper, we focus on the key flaw of the CSC model when adapting to multi-domain scenarios: the tendency to forget previously acquired knowledge upon learning new domain-specific knowledge (i.e., \textbf{catastrophic forgetting}).
To address this, we propose a novel model-agnostic \textbf{M}ulti-stage \textbf{K}nowledge \textbf{T}ransfer (\textbf{MKT}) framework, which utilizes a continuously evolving teacher model for knowledge transfer in each domain, rather than focusing solely on new domain knowledge. It deserves to be mentioned that we are the first to apply continual learning methods to the multi-domain CSC task. 
Experiments~\footnote{Our codes and data will be public after peer review.} prove the effectiveness of our proposed method, and further analyses demonstrate the importance of overcoming catastrophic forgetting for improving the model performance.
\end{abstract}

\section{Introduction}
\begin{CJK*}{UTF8}{gbsn}

Chinese Spelling Correction (CSC) plays a critical role in detecting and correcting spelling errors in Chinese text~\cite{li-etal-2022-past,ma-etal-2022-linguistic}, enhancing the accuracy of technologies like Optical Character Recognition (OCR) and Automatic Speech Recognition (ASR) ~\cite{afli-etal-2016-using,wang-etal-2018-hybrid}. In search engines, for example, CSC reduces human error, ensuring that users find the information they seek accurately.

\begin{table}[]
\setlength\tabcolsep{4pt} 
\small
\centering
\begin{tabular}{l|p{0.35\textwidth}}
\toprule
\midrule
\textbf{Input} &他通过了\textcolor{red}{张(zhāng)}基计划。\\
 & He passed the \textcolor{red}{Open} Foundation plan.\\
\midrule
\textbf{+EDU} &他通过了\textcolor{blue}{强 (qiáng)}基计划。\\
 & He passed the \textcolor{blue}{Strong} Foundation plan.\\ 
\textbf{+CHEM} &他通过了\textcolor{red}{羟 (qiǎng) }基计划。\\
 & He passed the \textcolor{red}{Hydroxyl} project.\\ 
\midrule
\textbf{Target} &他通过了\textcolor{blue}{强 (qiáng)}基计划。\\
 & He passed the \textcolor{blue}{Strong} Foundation plan.\\ 
\midrule
\bottomrule
\end{tabular}
\caption{Case of catastrophic forgetting in CSC.}
\label{tab1}
\end{table}

In practical applications, the input text may from various domains, demanding that the model contains different domain-specific knowledge. As illustrated in Table~\ref{tab1}, the word ``强基(Strong Foundation)'' is evidently common in Chinese Education domain. Accurately correcting ``张 (open)'' to ``强 (Strong)'' requires the model to have specific knowledge about the Chinese Education domain. Therefore, some related works begin to focus on the impact of multi-domain knowledge on the performance of CSC models~\cite{WuHongqiu2023}.

Previous works place greater emphasis on a model's ability to generalize to unseen domains, known as zero-shot performance, leveraging shared knowledge across different domains for generalization~\cite{liu2023chinese}. 
However, this paradigm falls short of enabling models to retain domain-specific knowledge, which is necessary for nuanced understanding and application.
Human learning processes can continuously acquire new domain-specific knowledge without losing their learned old knowledge.
Therefore, in this paper, we first investigate continual learning, which aligns perfectly with the human learning process, into CSC models for addressing this issue.

The core challenge of the continual learning setting is to minimize catastrophic forgetting of previously acquired knowledge while learning in new domains~\cite{wang2024comprehensive}. 
As demonstrated in Table~\ref{tab1}, when a CSC model learns the educational-specific word ``强基(Strong Foundation)'', it accurately corrects errors. However, after it continues to learn knowledge from the chemistry domain, it would learn the new knowledge of ``羟基(hydroxyl)'', but forget the education word ``强基(Strong Foundation)''.
However, in the previous works of multi-domain CSC, this catastrophic forgetting challenge remains unexplored.

To mitigate catastrophic forgetting in multi-domain CSC, 
we devise a multi-stage knowledge transfer framework based on continual learning, which employs a dynamically evolving teacher model that at each stage imparts all its previously accumulated knowledge to the current student model. Finally, through extensive experiments and analysis, we demonstrate the effectiveness of our proposed method. Our contributions are summarized as follows:

\begin{enumerate}
    \item We are the first to pay attention to the catastrophic forgetting phenomenon of multi-domain CSC, which is the key challenge that must be overcome for the CSC model to truly adapt to real multi-domain scenarios.
    \item We present a model-agnostic MKT framework that leverages the idea of continual learning to significantly suppress catastrophic forgetting.
    \item We conduct extensive experiments and solid analyses to verify the effectiveness and competitiveness of our proposed methods. 
\end{enumerate}

\end{CJK*}

\section{Our Approach}
Our approach is a special form of knowledge distillation that takes into account scenarios involving multiple stages of training. We leverage the knowledge acquired from these stages. This strategy of multi-stage knowledge transfer provides an effective solution to the challenges encountered in continual learning.

\subsection{Problem Formulation}
The CSC task is to detect and correct spelling errors in Chinese texts. Given a misspelled sentence $X = \{x_1,\ x_2,...,\ x_n\}$ with $n$ characters, a CSC model takes $X$ as input, detects possible spelling errors at character level, and outputs a corresponding correct sentence $Y = \{y_1,\  y_2,...,\ y_n\}$ of equal length. This task can be viewed as a conditional sequence generation problem that models the probability of $ p(Y|X)$.
In multi-domain CSC tasks, assuming that there are n domains $D=\{D_1,\  D_2,  ...,\ D_n\}$, these domains are trained sequentially, where each domain $D_k$ is trained without access to the data from previous domains, from $D_1$ to $D_{k-1}$. Furthermore, after training domain $D_k$, we should consider the performance of all domains from $D_1$ to $D_k$.

\begin{figure}[h]
    \centering
    \includegraphics[width=0.49\textwidth]{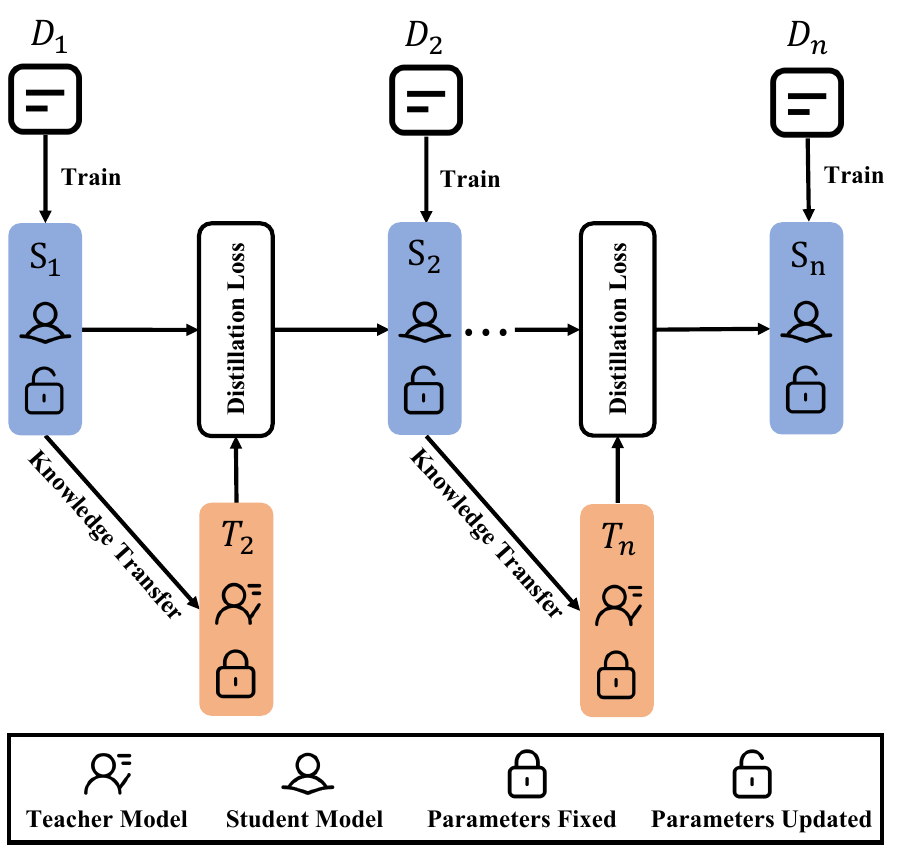}
    \caption{Overview of MKT framework.}
    \label{fig1}
\end{figure}

\subsection{Structure of MKT framework}

To tackle catastrophic forgetting, an intuitive solution is to transfer the knowledge previously acquired to the most recent model. The foundational idea revolves around transferring previously acquired knowledge to the latest model iteration. However, maintaining a distinct model for each stage quickly becomes untenable due to escalating storage and computational requirements with the addition of each stage.

To address this challenge, our framework employs a dynamic teacher model strategy. As illustrated in Figure~\ref{fig1}, this teacher model acts as a comprehensive knowledge repository, effectively serving as a backup of the student model from the previous stage to calculate the distillation loss for the current stage's student model. It encapsulates all the domain-specific knowledge accumulated to date, providing crucial guidance for the model training in the current phase.

\subsection{MKT framework for Multi-domain CSC}
We consider the scenario where the training is comprised of $m$ stages, denoted by $k = 1,2...,m$. At $k$-th stage, a subset of data $\{x^{(i)}_k,\  y^{(i)}_k \}^{T_k}_{i=1}$ are fed to the model, where $T_k$ refers to the number of samples at $k$-th stage, $x^{(i)}_k$ refers to $i$-th sample at $k$-th stage.

Assume that $u_k(\cdot)$ is an unknown target function that maps each $x^{(i)}_k$ to $y^{(i)}_k$ at stage $k$, i.e., $y^{(i)}_k$ = $u_k(x^{(i)}_k)$. Under the continual learning setting, our goal is to train a CSC model $g(\cdot\ ; w)$ parameterized by $w$, such that $g(\cdot\ ; w)$ not only fits well to $u_{k}(\cdot)$, but also fits $u_{k−1}(\cdot)\ , u_{k−2}(\cdot)$, · · ·  $,\ u_1(\cdot)$ in early stages to alleviate catastrophic forgetting.

\begin{table*}[ht]
\small
\centering\setlength\tabcolsep{12pt}
\begin{tabular}{llccccc}
\toprule
\midrule
\textbf{Backbone} & \textbf{Model}        & \textbf{General} & \textbf{CAR}  & \textbf{MED}   & \textbf{LAW}   & \textbf{Avg}    \\
\midrule
BERT     & Baseline    & 67.41   & 33.50 & 42.86 & 62.35 & 51.53 \\
         & +MKT(Ours)        & 67.90$^\uparrow$  & 35.86$^\uparrow$ & 43.46$^\uparrow$ & 62.88$^\uparrow$ & 52.53$^\uparrow$ \\
\midrule
                
Soft-Masked BERT  & Baseline	    & 54.22	& 30.73	& 43.88	& 68.54	& 49.34 \\
	          & +MKT(Ours)	      & 60.98$^\uparrow$  & 35.11$^\uparrow$ & 51.27$^\uparrow$ & 70.68$^\uparrow$ & 54.51$^\uparrow$ \\
\midrule
REALISE  & Baseline    & 70.78   & 27.48 & 53.33 & 70.59 & 55.55 \\
         & +MKT(Ours)	      & 72.74$^\uparrow$  & 29.25$^\uparrow$ & 55.28$^\uparrow$ & 70.85$^\uparrow$ & 57.03$^\uparrow$ \\
\midrule
\bottomrule
\end{tabular}
\caption{Performance on the test set of each domain after training on all datasets.}\label{table2}
\end{table*}

We need to minimize the loss function to optimize the model weights:
\begin{equation}\label{Eq.1}
L^{(k)}=\lambda L^{(k)}_s+(1-\lambda)L^{(k)}_h.
\end{equation}
In the equation, $\lambda$ is a hyper-parameter that ranges from $[0, 1]$. $L^{(k)}_s$ is the knowledge distillation loss, calculating cross entropy between the output probabilities of teacher model $g(\cdot\ ;w_{k-1})$ and student model $g(\cdot\ ;w_k)$:
\begin{equation}\label{Eq.2}
L^{(k)}_s=-\sum_{i=1}^{T_k} g(x_k^{(i)};\ \omega_{k-1})\times \log g(x_k^{(i)};\ \omega_k).
\end{equation}
$L^{(k)}_h$ is the cross-entropy loss between the output of student model $g(\cdot\ ; w_k)$ and ground truth $y_k$:
\begin{equation}\label{Eq.3}
L^{(k)}_h=-\sum_{i=1}^{T_k} y_k^{(i)} \times \log g(x_k^{(i)};\ \omega_k).
\end{equation}

\begin{algorithm}
\small
    \caption{\textbf{MKT Framework}}
    \label{alg:MKT}
    \begin{algorithmic}[1] 
        \Statex \textbf{Input:} Training set ${D_k}$, Student model ${S_{k-1}}$
        \Statex \textbf{Output:} Student model ${S_k}$
        \State Copy $S_{k-1}$ as the teacher model ${T_k}$
        \State Freeze the parameters of ${T_k}$
        \State ${S_k}$ forward propagation and calculates the loss guided by ${T_k}$ according to Equation \ref{Eq.1}
        \State Optimize the parameters of ${S_k}$
        \State \textbf{Return} ${S_k}$
    \end{algorithmic}
\end{algorithm}

As shown in Algorithm~\ref{alg:MKT}, during the training phase of the $k$-th domain, we employ the model refined from the preceding $k-1$ domains, i.e, $S_{k-1}$, as the teacher model $T_k$, alongside the concurrently trained student model $S_k$. The parameters of $T_k$ are frozen. 
The final loss is a weighted summation of the knowledge distillation loss $L^{(k)}_s$ and the original loss of the CSC task $L^{(k)}_h$.

\section{Experiment and Result}

\subsection{Datasets and Metrics}
Considering the multi-domain setting we focus on, we set up four domains, namely \textbf{General, Car, Medical, and Legal} domains.
The reason for this setting is that the differences in characteristics between these domains are the most obvious, which brings the most serious catastrophic forgetting to CSC models.
For the general domain, as in previous work, we also use SIGHAN13/14/15~\cite{wu-etal-2013-chinese,yu2014chinese,tseng-etal-2015-introduction} and Wang271K~\cite{wang-etal-2018-hybrid} as training data and SIGHAN15 test set as our test data.
For other special domains, we utilize the data resources released by LEMON~\cite{WuHongqiu2023} and ECSpell~\cite{Lv_2023}, and randomly take 500 samples from the original data of each domain as the test set. The dataset statistics are presented in the Appendix~\ref{appendix:dataset}.

Our evaluation predominantly relies on the sentence-level F1 score, a widely acknowledged metric. This criterion is notably stringent, adjudging a sentence as accurate solely when every error within is precisely identified and rectified, thereby providing a more rigorous evaluation compared to character-level metrics.

\subsection{Baseline Methods}

We select three widely used CSC baselines that embody varying integration of sensory inputs, to assess the efficacy of our method in diverse structural contexts: \textbf{\textit{BERT}}~\cite{devlin-etal-2019-bert} is to directly fine-tune the \textit{chinese-roberta-wwm-ext} model with the training data. \textbf{\textit{Soft-Masked BERT}}~\cite{zhang-etal-2020-spelling} incorporates a soft masking process after the detection phase, where it calculates the weighted sum of the input and [MASK] embeddings. \textbf{\textit{REALISE}}~\cite{xu-etal-2021-read} models semantic, phonetic and visual information of input characters, and selectively mixes information in these modalities to predict final corrections. Other implementation details are shown in the Appendix~\ref{appendix:implementation}.

\subsection{Results and Analyses}

\paragraph{Main Results} From Table~\ref{table2}, we see that after the optimization of our MKT, whether it is BERT, Soft-Masked BERT specially designed for CSC, or REALISE that integrates multi-modal information, performance improvements have been achieved in all domains. This reflects the effectiveness and the model-agnostic characteristic of our proposed MKT framework.

\begin{table}[ht]
\small
\centering
\setlength\tabcolsep{6pt} 
\begin{tabular}{l|ccccc}
\toprule
\midrule
\textbf{\(\lambda\)}  & \textbf{General} & \textbf{CAR}  & \textbf{MED}   & \textbf{LAW}   & \textbf{Avg }
\\
\midrule
0       & 67.41 & 33.50	& 42.86	& 62.35	& 51.53 \\
0.001	& 65.42	& 34.00$^\uparrow$	& 43.13$^\uparrow$	& 62.07	& 51.16 \\
\midrule
0.005	& 65.92	& 34.80$^\uparrow$	& 43.19$^\uparrow$	& 62.22	& 51.53$^\uparrow$ \\
0.01	& 67.90$^\uparrow$	& 35.86$^\uparrow$	& 43.46$^\uparrow$	& 62.88$^\uparrow$	& 52.53$^\uparrow$ \\
0.015	& 66.73	& 36.10$^\uparrow$	& 44.29$^\uparrow$	& 61.97	& 52.27$^\uparrow$ \\
0.02	& 66.48	& 37.47$^\uparrow$	& 42.92$^\uparrow$	& 62.63$^\uparrow$	& 52.38$^\uparrow$ \\
\midrule
0.05	& 67.41	& 32.32	& 41.98	& 61.66	& 50.84 \\
0.1	    & 68.18$^\uparrow$	& 31.02	& 41.09	& 60.62	& 50.23 \\
0.2	    & 67.74$^\uparrow$	& 27.55	& 38.31	& 56.96	& 47.64 \\
0.5	    & 66.47	& 23.53	& 27.09	& 44.44	& 40.38 \\
0.8	    & 60.64	& 12.35	& 15.15	& 26.22	& 28.59 \\
\midrule
\bottomrule
\end{tabular}
\caption{Performance of BERT+MKT on each domain after training across all domains with different \(\lambda\).}\label{table3}
\end{table}

\begin{figure}[h]
    \centering
    \includegraphics[width=0.32\textwidth]{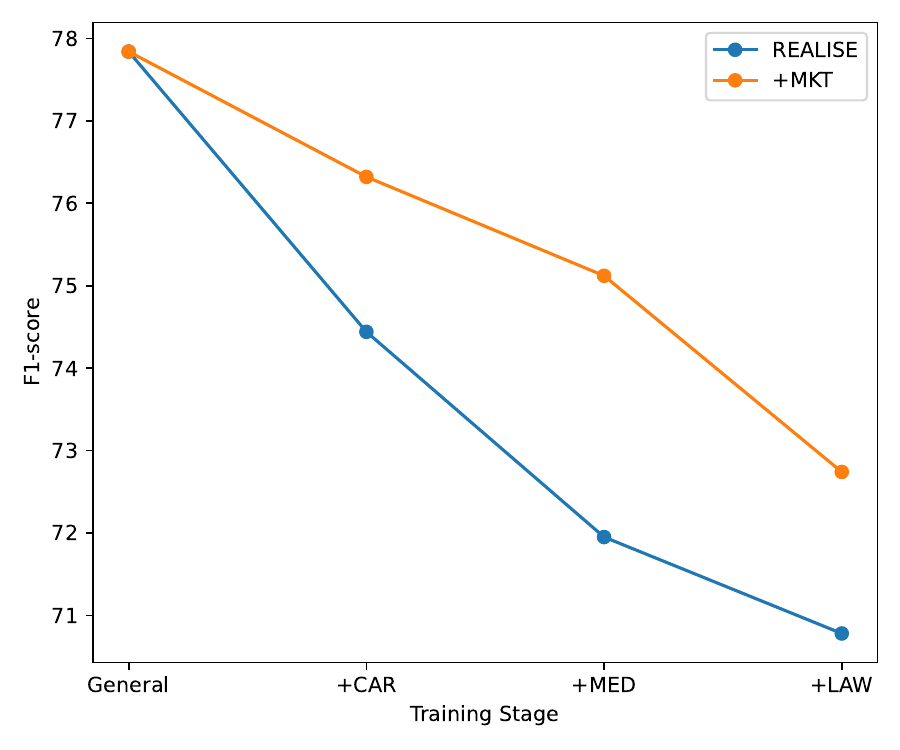}
    \caption{The phenomenon of model forgetting general-domain knowledge during incremental domain training.}
    \label{fig2}
\end{figure}

\paragraph{Parameter Study}
To explore the impact of the key parameter $\lambda$,  we conduct experiments on BERT+MKT using varying $\lambda$ values. As Table~\ref{table3} indicates, settings $\lambda$ between 0.005 and 0.02 stably bring improvements over the baseline. 
Particularly, setting $\lambda$ at 0.01 performs best in all domains.
We think that the main reason for this phenomenon is that the amount of training data in each special domain accounts for approximately 1\% of the amount of general training data (as shown in Appendix~\ref{appendix:dataset}).
Therefore, intuitively for MKT, an appropriate $\lambda$ can be selected based on the ratio of general training data to training data in other domains to obtain optimal performance.

\paragraph{Catastrophic Forgetting}
As shown in Figure~\ref{fig2}, we select the best-performing model (i.e., REALISE) in Table~\ref{table2} to observe its performance loss (i.e. catastrophic forgetting) in the general domain after being incrementally trained with other domain data.
Obviously, we see that after the optimization of MKT, the performance loss of REALISE is much smoother, which shows that catastrophic forgetting is well alleviated by our proposed MKT.

\subsection{Case Study}
\begin{CJK*}{UTF8}{gbsn}

\begin{table}[ht]
\setlength\tabcolsep{4pt} %
\small
\centering
\begin{tabular}{lr}
\toprule
\midrule
\multicolumn{2}{l}{\textbf{Circumventing Catastrophic Forgetting}} \\
\midrule
\textbf{Input}         &  年轻人的\textcolor{red}{青}量级玩乐SUV\\
\midrule
\textbf{+CAR(REALISE)}   &  年轻人的\textcolor{blue}{轻}量级玩乐SUV\\
\textbf{+CAR(+MKT)}       &  年轻人的\textcolor{blue}{轻}量级玩乐SUV\\
\textbf{+MED(REALISE)}   &  年轻人的\textcolor{red}{氰}量级玩乐SUV\\
\textbf{+MED(+MKT)}       & 年轻人的\textcolor{blue}{轻}量级玩乐SUV\\
\midrule
\textbf{Target} &年轻人的\textcolor{blue}{轻}量级玩乐SUV\\ \midrule
\bottomrule
\end{tabular}
\caption{Cases from the CAR test set to show MKT mitigates over-correction and catastrophic forgetting.}
\label{your-label}
\end{table}

To further verify the effectiveness of our MKT in mitigating catastrophic forgetting in multi-domain CSC, we present some cases in Table~\ref{your-label}.
As shown in table~\ref{your-label}, for the test sentence in the CAR domain, when REALISE has just been trained on the CAR domain, it can accurately correct errors. However, when REALISE is then trained on the MED domain, it can no longer correct successfully and instead predicts ``氰(cyanide)'' related to the medical domain. This is a typical catastrophic forgetting case where old domain knowledge is washed away by new domain knowledge. It can be seen that with the optimization of MKT, REALISE effectively avoids the occurrence of catastrophic forgetting.

\end{CJK*}

\section{Conclusion}
This paper demonstrates through experimentation that existing CSC models, when adapting to multi-domain scenarios, tend to forget previously acquired knowledge while learning new domain-specific information, a phenomenon known as catastrophic forgetting. Consequently, we propose an effective, model-agnostic framework for multi-stage knowledge transfer to mitigate catastrophic forgetting. Extensive experiments and detailed analyses demonstrate the importance of catastrophic forgetting we focus on and the effectiveness of our proposed method.

\clearpage
\section*{Limitations}

We do not compare our proposed method against commonly used Large Language Models (LLMs) in our experiments. The primary reason is that in the CSC task, representative LLMs still lag behind traditional fine-tuned smaller models, which has been proved by many related works. 
In addition, our approach specifically focuses on the Chinese scenarios. However, other languages, such as English, could also benefit from our methodology. We will conduct related studies on English scenarios in the future.



\bibliography{anthology,custom}
\bibliographystyle{acl_natbib}

\appendix


\section{Related Work}
\label{appendix:related_work}

This section comprehensively reviews CSC research, structured according to the data flow within correction models, and also delves into three principal methods in continual learning.

\subsection{Chinese Spelling Correction}
In CSC, we witness significant advancements in various model architectures and modules~\cite{DBLP:conf/emnlp/LiMZLLHLLC022, DBLP:journals/corr/abs-2307-09007, DBLP:conf/icassp/ZhangLZMLCZ23, DBLP:conf/emnlp/YeLZLM0023, DBLP:journals/corr/abs-2210-12692, DBLP:journals/corr/abs-2306-17447, DBLP:conf/emnlp/YeLL023, DBLP:conf/emnlp/HuangYZLLZZ23, DBLP:journals/corr/abs-2311-11268}. Early models like Confusionset-guided Pointer Networks optimize at the dataset level, leveraging confusion sets for character generation to enhance accuracy through commonly confused characters~\cite{wang-etal-2019-confusionset}. Innovations in embeddings, such as REALISE, improve model inputs by integrating semantic, phonetic, and visual information into character embeddings~\cite{xu-etal-2021-read}. Encoder improvements are highlighted by Soft-Masked BERT, which employs Soft MASK techniques post-detection to blend input with [MASK] embeddings for effective error prediction~\cite{zhang-etal-2020-spelling}. SpellGCN innovatively constructs a character graph, mapping it to interdependent detection classifiers based on BERT-extracted representations~\cite{cheng-etal-2020-spellgcn}. While previous multi-domain CSC research emphasizes shared knowledge and generalization across domains, this paper pioneers in addressing the catastrophic forgetting of domain-specific knowledge.

\subsection{Continual Learning}
In continual learning, replay, regularization, and parameter isolation stand as core strategies~\cite{DBLP:journals/patterns/LiuLTLZ22, DBLP:conf/sigir/LiLHYS022, Wang2023ACS, DBLP:journals/csur/DongLGCLSY23, DBLP:journals/tkde/LiHZZLLCZS23}. Replay methods like GEM and MER retain training samples, using constraints or meta-learning to align gradients~\cite{NIPS2017_f8752278,Riemer2018LearningTL}. Regularization, exemplified by Elastic Weight Consolidation (EWC), focuses on preserving task-specific knowledge by prioritizing parameter importance~\cite{doi:10.1073/pnas.1611835114}. Knowledge distillation aims at incremental training, transferring insights from larger to smaller models~\cite{GouJianping2021}. Parameter isolation techniques, such as CL-plugin, allocate unique parameters to different tasks, reducing interference~\cite{ke-etal-2022-continual}. Our work introduces continual learning to multi-domain CSC for the first time, with our MKT framework being model-agnostic across various CSC models.

\section{Statistics of the datasets}

\label{appendix:dataset}
\begin{table}[ht]
\small
\setlength\tabcolsep{6pt} 
\begin{tabular}{llrrr}
\toprule
\midrule
Training Set & Domain  & Sent & Avg.Length  & Errors   \\
\midrule
Wang271K & General      & 271,329 & 42.6	& 381,962 \\
SIGHAN13 & General     & 700     & 41.8	& 343   \\
SIGHAN14 & General     & 3,437   & 49.6	& 5,122 \\
SIGHAN15 & General     & 2,338   & 31.1	& 3,037 \\
CAR  & CAR	      & 2,744	& 43.4	& 1,628 \\
MED & MED	      & 3,000	& 50.2	& 2,260 \\
LAW & LAW	      & 1,960	& 30.7	& 1,681	 \\
\midrule
\midrule
Test Set & Domain  & Sent & Avg.Length & Errors   \\
\midrule
SIGHAN15 & General     & 1,100   & 30.6	& 703 \\
CAR  & CAR	      & 500 	& 43.7	& 281 \\
MED & MED	      & 500 	& 49.6	& 356 \\
LAW & LAW	      & 500 	& 29.7	& 390 \\
\midrule
\bottomrule
\end{tabular}
\caption{Statistics of the datasets, including the number of sentences, the average length of sentences in tokens, and the number of errors in characters.}\label{table5}
\end{table}

\section{Implementation Details}
\label{appendix:implementation}
For the three models, we initially train them on the general dataset, followed by successive training on the CAR, MED and LAW datasets. After the completion of training, we test the final models' performance on all datasets.

In the experiments, we train on the aforementioned datasets for 10 epochs each, with a batch size of 64 and a learning rate of 5e-5. The baseline method simply involved sequential training on the same model across the mentioned datasets. Our approach, however, included a knowledge transfer process in each phase, where the $\lambda$ between $L_h$ and $L_s$ was set to 0.01.

\end{document}